\documentclass{llncs}

\usepackage{graphicx} 
\usepackage{subfigure} 

\usepackage{hyperref}

\usepackage{amsmath,amssymb}
\usepackage{booktabs}
\usepackage{multirow}

\usepackage{xspace}

\newcommand{\mat}[1]{\mathbf{#1}}

\newcommand\sectionname{Sect.}
\newcommand\eg{\textit{e.g.},\xspace}
\newcommand\ie{\textit{i.e.},\xspace}

\newcommand\dimension{D}

\usepackage{upgreek}

\newcommand\identityMatrix[1]{\mat{I}}

\usepackage{color}

\definecolor{myOrange}{RGB}{192,96,0}

\newcommand\myparagraph[1]{\vspace{8pt}\noindent\textbf{#1}\quad}

\usepackage{fancyhdr}
\begin{document} 

\begin{center}
\Huge 
\bf
\null
\vfill
\vspace{-4cm}
Technical Report%
\\
\vspace{2cm}
TR-FSU-INF-CV-2014-01%
\vfill
\clearpage
\end{center}

\newpage

\title{Seeing through bag-of-visual-word glasses: towards understanding quantization effects in feature extraction methods}
\newcommand\myShortAuthors{A. Freytag, J. R\"uhle, P. Bodesheim, E. Rodner, and J. Denzler}
\author{Alexander Freytag, Johannes R\"uhle, Paul Bodesheim, Erik Rodner, and Joachim Denzler}
\newcommand\myShortTitle{Seeing through bag-of-visual-word glasses}

\institute{Computer Vision Group, Friedrich Schiller University Jena, Germany\\\url{http://www.inf-cv.uni-jena.de}}

\maketitle
\setcounter{page}{1}

\begin{abstract}
Vector-quantized local features frequently used in bag-of-visual-words approaches are the backbone of popular visual recognition systems due to both their simplicity and their performance. 
Despite their success, bag-of-words-histograms basically contain low-level image statistics (e.g., number of edges of different orientations). 
The question remains how much visual information is “lost in quantization” when mapping visual features to code words?
To answer this question, we present an in-depth analysis of the effect of local feature quantization on human recognition performance. 
Our analysis is based on recovering the visual information by inverting quantized local features and presenting these visualizations with different codebook sizes to human observers.
Although feature inversion techniques are around for quite a while, to the best of our knowledge, our technique is the first visualizing especially the effect of feature quantization. 
Thereby, we are now able to compare single steps in common image classification pipelines to human counterparts\footnote{An abstract version of this paper was accepted for the ICPR FEAST Workshop.}. 
\end{abstract}

\fancyhead[LE]{\myShortTitle}
\fancyhead[RO]{\myShortAuthors}
\cfoot{\textsf{\thepage}}

\section{Introduction}
\label{sec:intro}

Traditionally, standard image classification systems follow a typical architecture: (1) pre-processing, (2) feature extraction, and (3) training and classification. 
A significant number of  current image categorization methods still follows the bag-of-visual-words (BoW) approach for feature extraction:
local features on a dense grid (\eg SIFT) are extracted and grouped by (un)supervised clustering for codebook creation (\eg k-Means), which then allows for
assigning local features to groups and for forming histograms that can be used as image representations~\cite{csurka2004visual,Lazebnik06:BBF,wang2010locality,Kapoor10:GP,Vedaldi12:EAK,Rodner12:LGP}.
The popularity of the BoW strategy is also apparent when looking at the list of Pascal VOC submissions~\cite{Everingham10:VOC}, where the majority of recognition systems can be perfectly mapped to the above
outlined pipeline. 

Clustering of local features has most often been motivated by the analogy of words for text categorization~\cite{csurka2004visual}. 
However, the discovered clusters usually correspond to blob-like objects being semantically poor, and the power of resulting image representations stems from informative statistics rather than from interpretable semantic parts.

\begin{figure} [tb]
  \centering
  \includegraphics[width=0.55\linewidth]{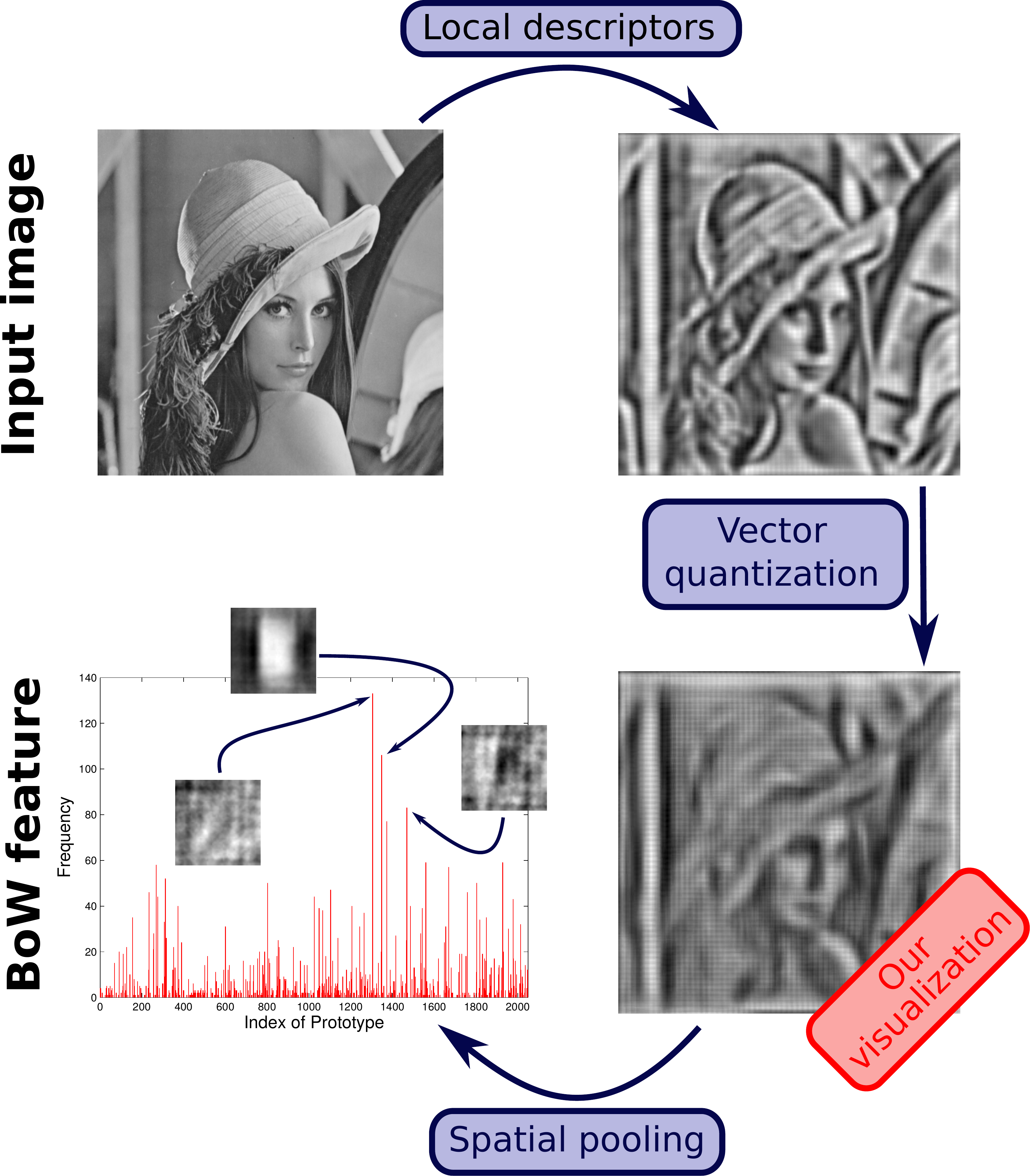}
  \caption{The long way from an image to its bag-of-visual-words representation. In this paper, we aim at visualizing the loss of information during vector quantization.
          }
  \label{fig:fromImgToBow}
\end{figure}

In the following paper, we like to see behind the BoW curtain by inspecting how much information is usually lost in vector quantizing local features to pre-computed codebooks.
We believe that our analysis is valuable for researchers trying to improve BoW models as well as for developers who try to build a
good image recognition system.
Therefore, we present a new method that inverts quantized local features and visualizes the information loss during this process (see \figurename~\ref{fig:fromImgToBow} for a vizualisation thereof).
Our inversion method is easy to implement and builds upon the work of \cite{Vondrick13:HOG}, where histogram of oriented gradient (HOG) features are inverted to study 
the visual information loss occurring during feature extraction. In contrast to previous work in this area~\cite{Weinzaepfel11:RAI,Vondrick13:HOG}, 
we focus on the effects of vector-quantization within the BoW model and
to our knowledge, we are the first qualitatively and quantitatively studying this aspect by asking human observers about estimated image content after inversion.

\section{Related work}
\label{sec:relatedWork}

\myparagraph{A brief history of bag-of-visual-words}
The bag-of-visual-words model goes back to \cite{csurka2004visual}, where it was first shown that histograms built on vector quantized local features
are highly suitable for object recognition tasks. Forming a single histogram by spatially pooling quantized features over the whole image and discarding any spatial information
was regarded as efficient option for obtaining occlusion-invariant features.
It was the common understanding at this time that interest point detectors are necessary to select
only local features at certain positions which are also suitable for image matching.
Later on, \cite{Nowak06:SSB} showed that this is not the case and 
that dense and random selection of local features allows for larger BoW descriptors and also for higher recognition rates. 
The evaluation paper of \cite{zhang2007local} further studied the influence of detector invariance properties and
demonstrated that, for example, invariance with respect to rotations and affine transformations explicitly hurts recognition performance.

The authors of \cite{grauman2005pyramid} and \cite{Lazebnik06:BBF} showed how pyramid matching and simple spatial pooling allows further performance boosting by re-incorporating
rough spatial information of local features.
The importance of a proper feature encoding with a given codebook was highlighted in \cite{coates2011importance}, where it was shown that the actual choice of cluster
method does not influence recognition results significantly and even a random clustering is sufficient.
The most important contributions in the area of feature encoding are the work of \cite{deselaers2005improving}, where soft quantization was first proposed, as well as the paper of \cite{fisher}, where
the authors developed an encoding method based on Fisher vectors, and the locality-constrained linear coding (LLC) method of \cite{wang2010locality}. The key idea of the LLC method is 
to use only the $k$ nearest neighbors in a codebook for encoding.

\myparagraph{Image reconstruction from local features}
Reconstructing an image from a given feature recently gained attention within our community to better understand learned models. 
In one of the first works within this area, the authors of \cite{Weinzaepfel11:RAI} propose a technique to reconstruct an image from densely extracted SIFT descriptors. 
How to invert local binary patterns used for the task of face identification was introduced by \cite{Angelo12:FBI}. 
Noteworthy, the authors have not been directly interested in inspecting feature capacity or learned models, but pointed to the problem that local 
features still contain lots of visual information and thus can be critical from a juristic point of view for face identification systems.
A visualization technique for popular HOG features was given in \cite{Vondrick13:HOG} which allowed for insights why object detectors sometimes fire at unexpected positions.
The work most similar to the current paper was recently published in \cite{Kato14:IRB}, which aims at inverting a given bag-of-visual-word histogram by first
randomly arranging prototypes and then optimize their positions based on adjacency costs. 
Since \cite{Kato14:IRB} measures inversion quality only in terms of reconstruction error to the original images,
it would be interesting to combine their inversion technique and our evaluation method based on asking human observers.

\myparagraph{Outline of this paper}
The remainder of the paper is structured as follows: 
Our simple yet insightful inversion technique is presented in \sectionname~\ref{sec:hogglebow}. 
We then provide a detailed comparison between machine learning performance and human performance for the task of
scene categorization in \sectionname~\ref{sec:experiments}, and present our webpage for accessing the evaluation server and participating in the large-scale study.
A summary of our findings in \sectionname~\ref{sec:conclusions} conclude the paper.

\section{Unbagging bag-of-visual words: visualizing quantization effects}
\label{sec:hogglebow}

Our technique is simple and in line with current trends for image reconstruction from local features~\cite{Weinzaepfel11:RAI,Angelo12:FBI,Vondrick13:HOG}. For an unseen image, 
we extract local features on a dense grid and follow the bag-of-words paradigm by quantizing them using a pre-computed codebook.
Based on inversion techniques for local features (see~\cite{Weinzaepfel11:RAI,Angelo12:FBI,Vondrick13:HOG} for impressive results),
we can compute the most probable image patch for every prototype, \ie we can visually inspect the quantization quality for a given codebook.
Thus, for any local feature $\boldmath{x}$, we vector-quantize it with a codebook and 
draw the inverted prototype into the reconstruction image 
with position and size according to the support of the extracted local feature. 
The complete pipeline of BoW-computation is visualized in \figurename~\ref{fig:fromImgToBow}.
In contrast to previous works for feature inversion,
which aim at inspecting the image in the top-right corner, out techniques is designed for inspecting the following step 
and consequently aims at visualizing quantization effects.
It should be noted that for the simplicity of demonstration, we chose HOG-features where inversion was successfully presented in \cite{Vondrick13:HOG} and source code is publicly available.
However, our method is not restricted to HOG-features and can be applied to any local feature type where inversion techniques are known for (\eg \cite{Weinzaepfel11:RAI,Angelo12:FBI}).
Our source code is available at \url{http://www.inf-cv.uni-jena.de/en/image_representation}.

\newcommand\px{\operatorname{px}}
\begin{figure}[bt]  \centering  
	\includegraphics[width=0.23\linewidth]{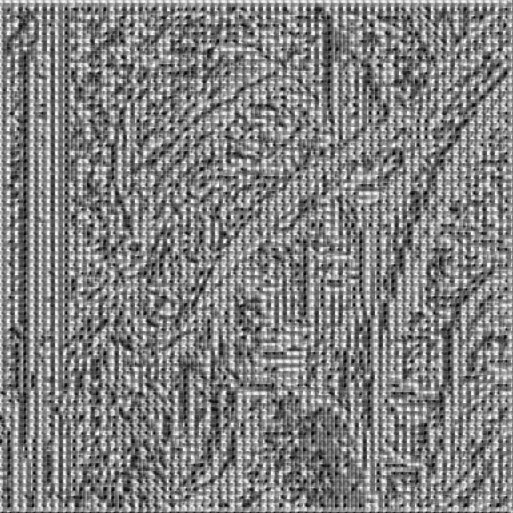}
	\includegraphics[width=0.23\linewidth]{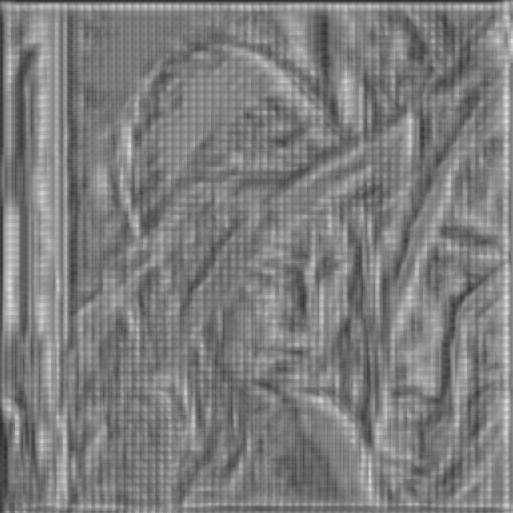}
	\includegraphics[width=0.23\linewidth]{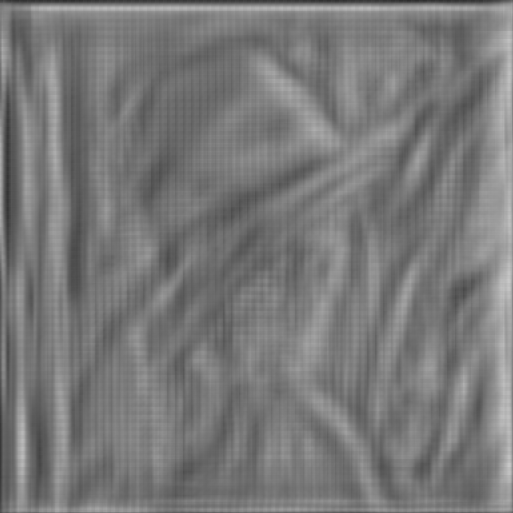}
	\includegraphics[width=0.23\linewidth]{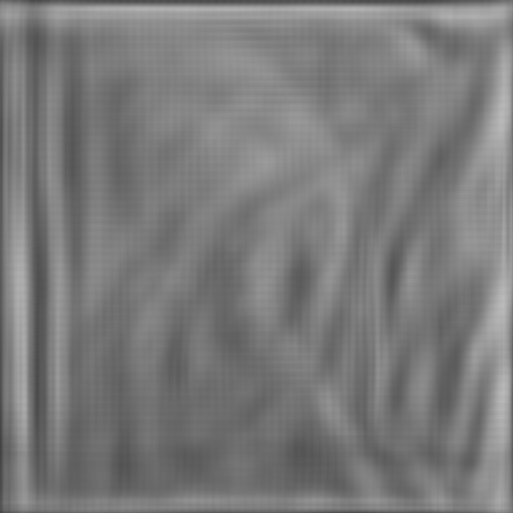}
  \caption{How does the size of extracted local features influence the visual quality when using our inversion technique? From left to right: patch sizes are $16\px$, $32\px$, $64\px$, and $128\px$, respectively.}
  \label{fig:patchSizes}
\end{figure}

\myparagraph{Effect of local patch sizes}
With the inversion method at hand, we can investigate the dependencies between several variables during feature extraction and the resulting visual quality.
Let us first look at the effect of feature extraction support, \ie sizes of patches local features are extracted from, on the resulting visual quality when using our inversion technique.
In \figurename~\ref{fig:patchSizes}, inversion results are displayed for increasing patch sizes for local feature extraction. For region sizes too small, 
extracted features can hardly capture any high-level statistics, and thus the resulting inversion looks heavily 'cornered'. On the other hand,
for regions too large, extracted local features are highly diverse, and negative aspects of quantization become visible. 

\begin{figure}[btp]  \centering  
	\includegraphics[width=0.23\linewidth]{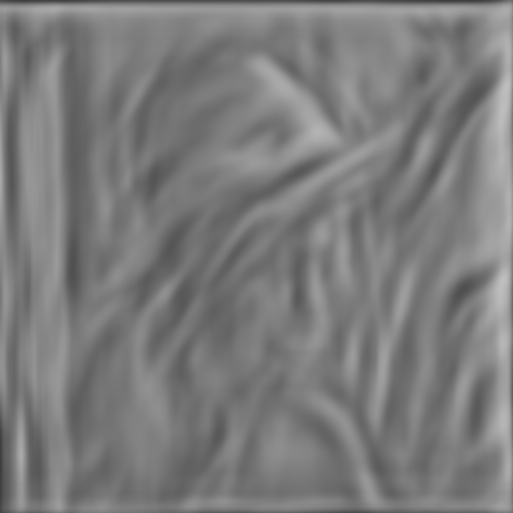}
	\includegraphics[width=0.23\linewidth]{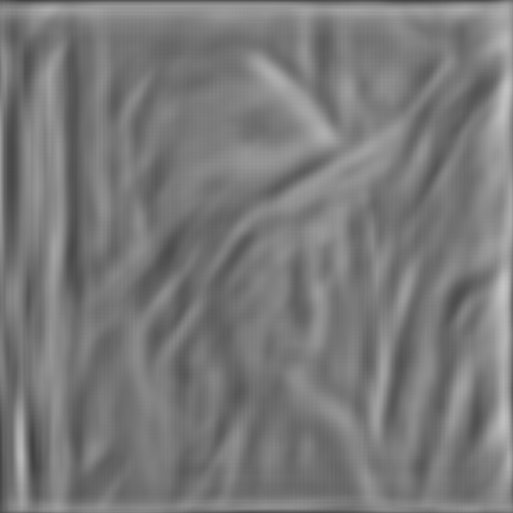}
	\includegraphics[width=0.23\linewidth]{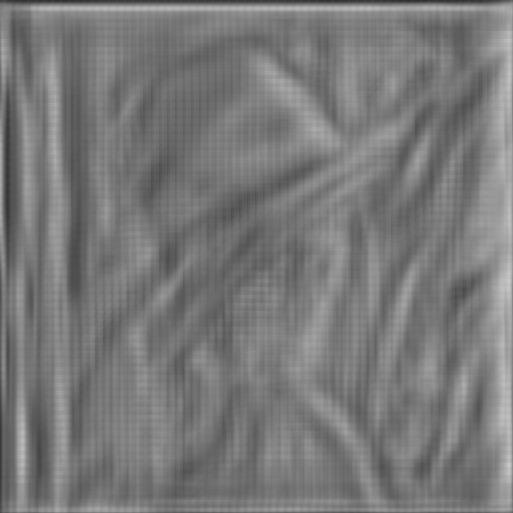}
	\includegraphics[width=0.23\linewidth]{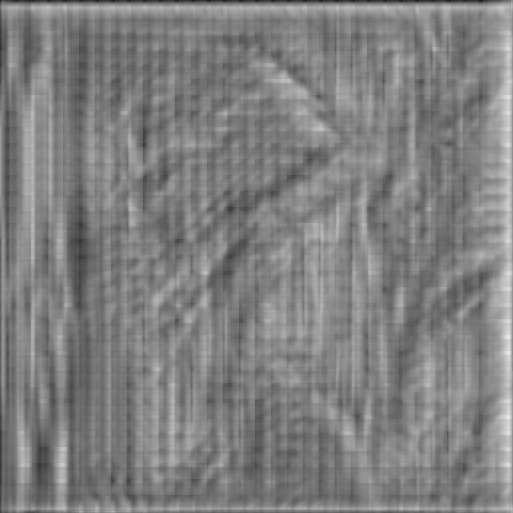}
  \caption{How does the stride width for local feature extraction influence the visual quality when using our inversion technique? From left to right: stride of $2\px$, $4\px$, $8\px$, and $16\px$, respectively.}
  \label{fig:stride}
\end{figure}

\myparagraph{Effect of patch stride}
As a second aspect, we investigate the effect of stride width during local feature extraction on visual quality of reconstructed images.
For different strides between $2\px$ and $16\px$, reconstruction results are given in \figurename~\ref{fig:stride}.
As can be seen, small stride widths tend to average out high-frequency parts, whereas higher strides result in edge artifacts at the boundaries of extracted local patches.

In summary, depending on the local features and inversion method at hand, we can easily inspect dependencies of parameter settings on the visual quality.
Thus, we can implicitly estimate which parameter configuration preservers or neglects certain kinds of information present in original images.

\section{Experimental evaluation}
\label{sec:experiments}

Our experiments are based on a scene classification task and in particular, we use the $15$ Scenes dataset of \cite{oliva2001modeling}. This task was chosen, because
it is also difficult for human observers due to the moderate number of classes (see human performance on original images in \sectionname~\ref{subsec:exp})
and the fine-grained details that are necessary to distinguish between different scene categories, \eg street \textit{vs.} highway.

\subsection{Machine learning baseline}

Local features are extracted from overlapping $64\times64$ image patches on a dense grid with a stride of $8$ pixel and zero padding on image borders.  
As underlying representations, we choose the commonly used variant of HOG-features as presented in~\cite{Felzenszwalb10:ODD}. In general, $4\times4$ HOG blocks are computed resulting
in $\dimension = 512$ dimensional features. Clustering is done using the k-Means implementation of VLFeat~\cite{Vedaldi08:vlfeat}.
All classes are learned with $100$ images during training for the $15$ Scenes dataset.
Classification is performed using LibLinear~\cite{Fan08:liblinear} and explicit kernel maps~\cite{Vedaldi12:EAK} to increase model complexity ($\chi^2$\!-approximation with $n=3$ and $\gamma=0.5$
as suggested by~\cite{Vedaldi12:EAK}).
All classification results presented are averaged over $25$ random data splits.
Regularization parameters are optimized using $10$-fold cross-validation. 
For further details, we point to the source code 
released on the project page\footnote{\url{http://www.inf-cv.uni-jena.de/en/image_representation}}. 

\subsection{Experimental setup for human experiments}
\label{subsec:exp}

\begin{figure}[tb]
  \centering  
  \includegraphics[width=1\linewidth]{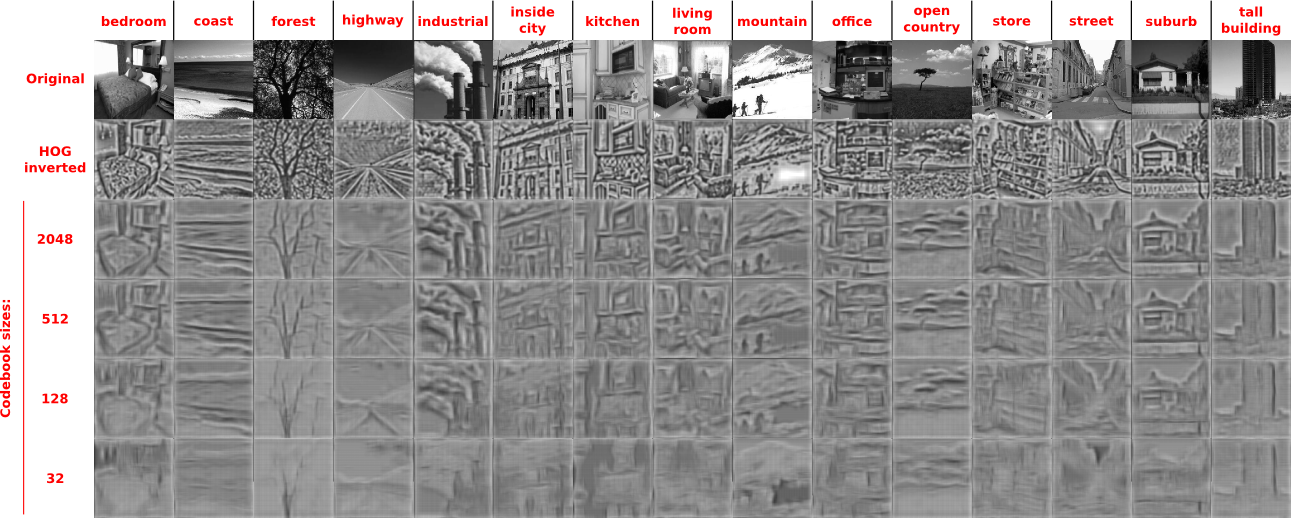}
  \caption{Overview of the images presented to human observers during the experiment (15 Scenes dataset).}
   \label{fig:experimentdata}
\end{figure}

Since the scene recognition task was unknown to most of our human observers, we showed them example images for each category in the beginning similar to \figurename~\ref{fig:webexperiment_screenshot}. 
Afterward, human subjects needed to classify new images and we randomly sampled visualizations with different 
quantization levels (original image, inverted HOG image without quantization, inverted HOG image with codebook size $k$; $k \in \{32,128,512,2048\}$). 
There was no time restriction during the test phase and human observers were allowed to see example images of the categories throughout the whole experiment. 
In total, we had 20 participants in our study by the time this paper was written, and most of them were colleagues from our group. 
Note that this can of course not be considered as a representative group of human subjects and results will be definitely biased.
However, the conclusions we can 
draw from this limited amount of data are still interesting and a large-scale study is currently running.

\subsection{Evaluation: are we lost in quantization?}

\begin{figure}[tb]
  \centering  
  \includegraphics[height=0.44\linewidth]{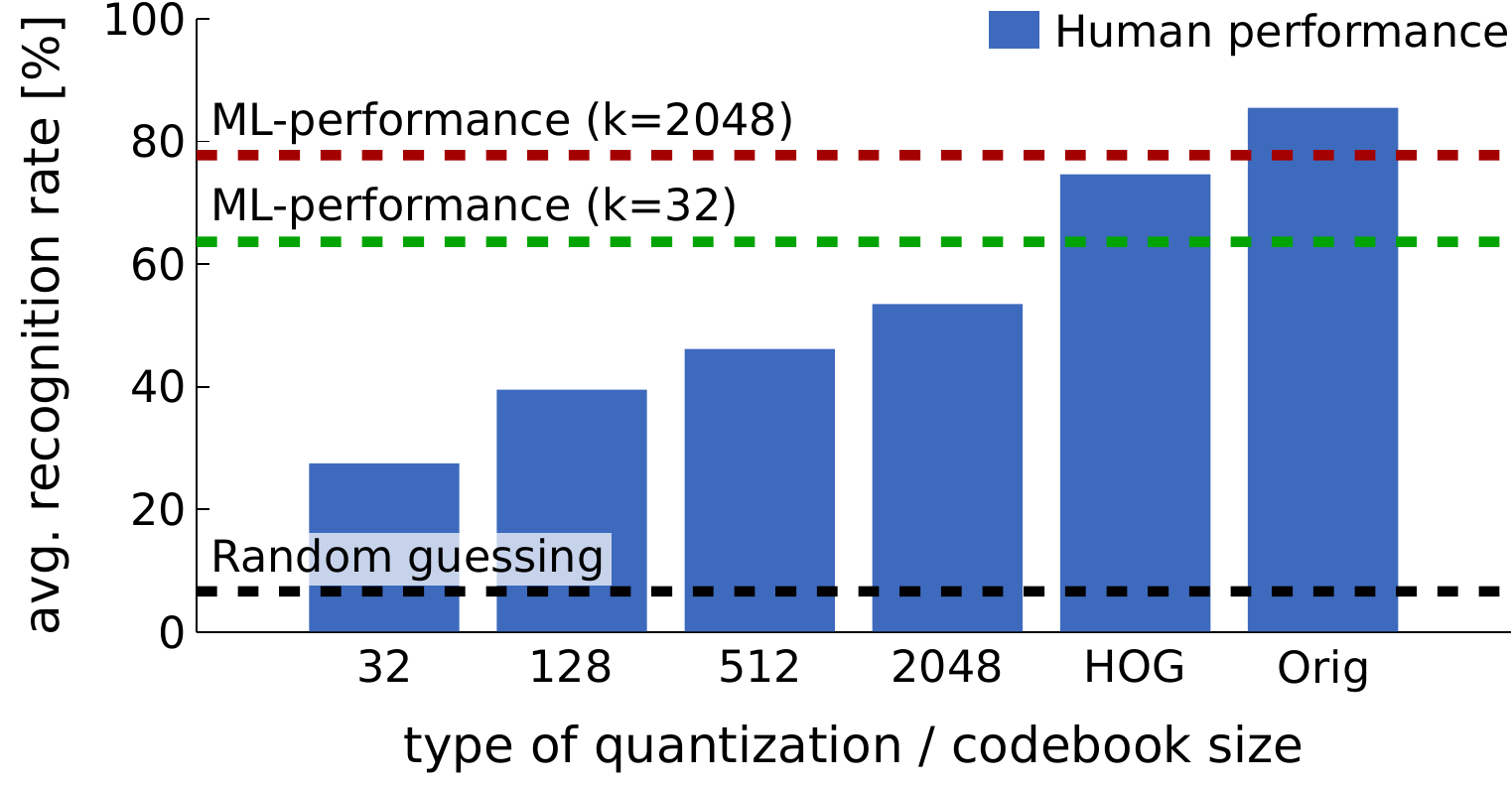}
  \caption{Human classification result in comparison to machine performance.}
  \label{fig:results}
\end{figure}

The main results of our experiments obtained by humans as well as machine learning techniques are given in \figurename~\ref{fig:results}. The plot shows the human scene recognition performance measured in terms of the average recognition rate depending on the type of quantization (original image, no quantization, quantization with a specific size of the codebook). 

As can be seen, the human recognition performance increases with the number of codebook elements, which is not a surprising fact. However, it is surprising that for a codebook size of $32$ human performance is significantly worse than machine learning performance (marked with a green line in the plot). This gap becomes  smaller when we increase the codebook size but it is still existing for $k=2,\!048$ and even when no quantization is used at all. Only when the original images are shown to human subjects, the machine learning bag-of-visual words method is not able to beat human performance.
It has to be noted here that the small gap between human and machine performance in this case is still surprising given the fact that the machine learning method is not provided with any spatial information.

\subsection{Web interface to the evaluation server}

\begin{figure}[tb]
  \centering  
  \includegraphics[height=0.32\textwidth]{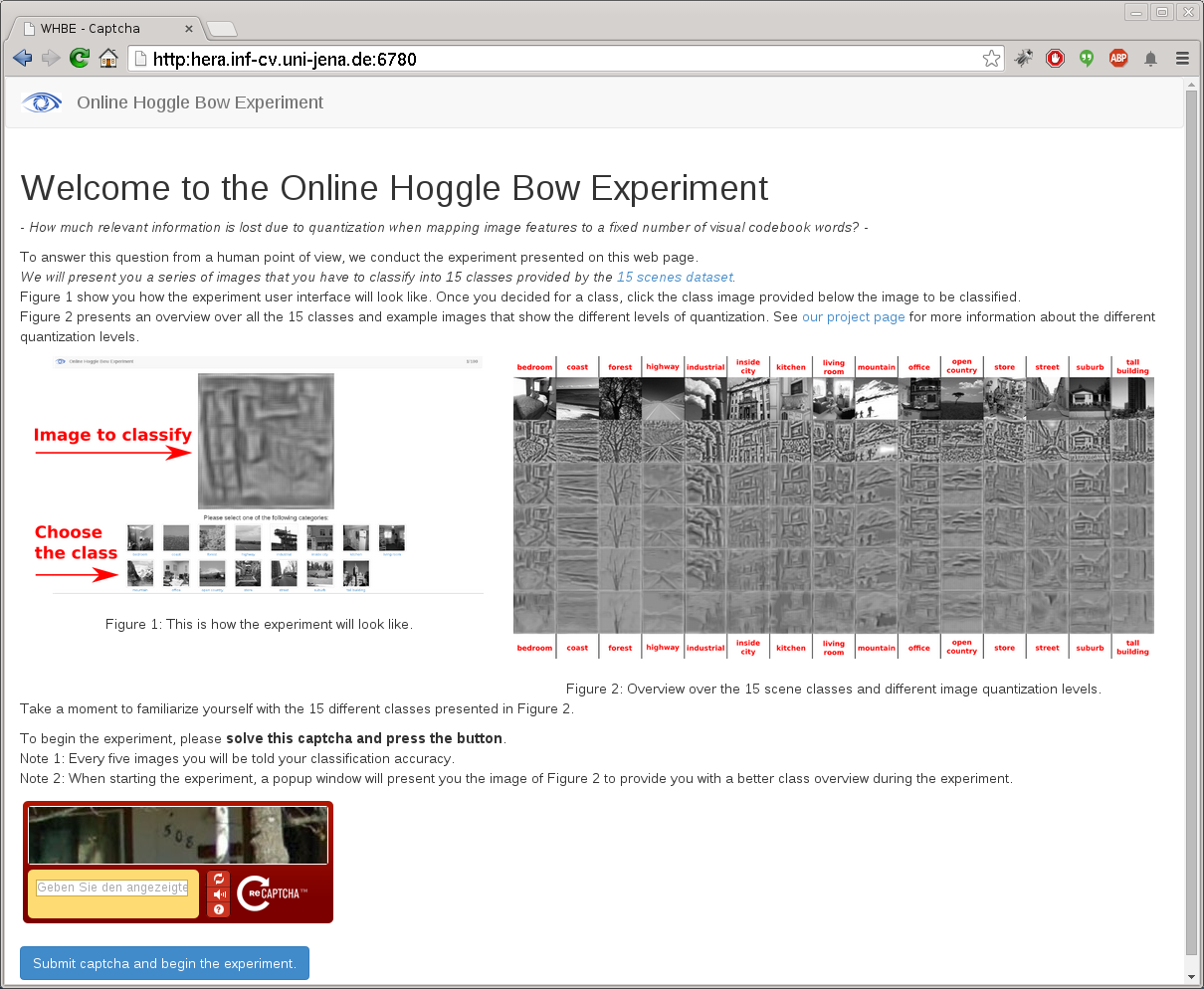}
  \includegraphics[height=0.32\textwidth]{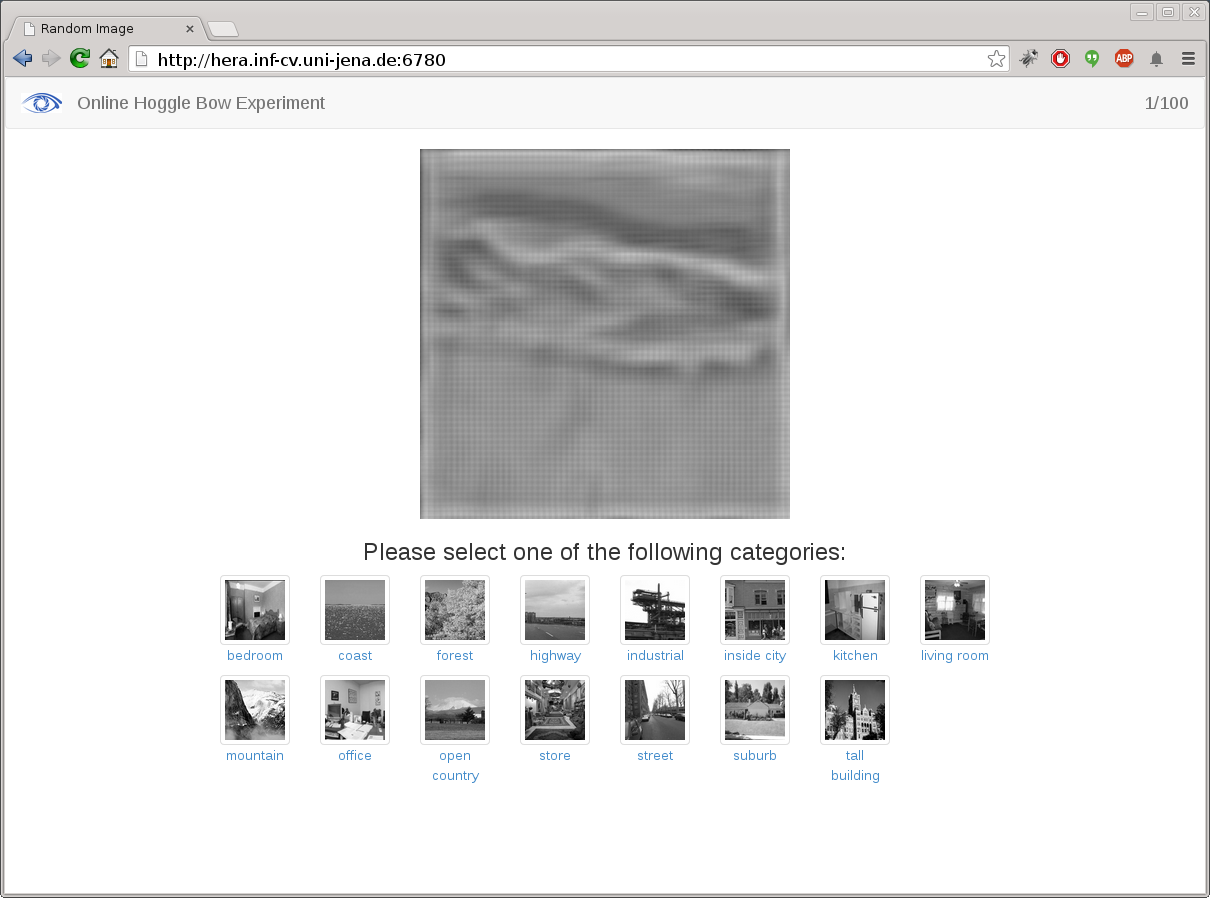}
  \caption{Screenshots showing the web experiment. After an introduction (\textit{left image}), the user has to classify the presented inverted image into the given 15 scenes (\textit{right image}).}
  \label{fig:webexperiment_screenshot}
\end{figure}

Our current results are based on only a small set of human subjects. However, we already prepared a large-scale web-based study and a corresponding web interface\footnote{The authors would like to acknowledge Clemens-Alexander Brust for writing an excellent \texttt{flask} application for the web interface.} for our human studies. Some screenshots are displayed in \ref{fig:webexperiment_screenshot}. 
The web interface can be access under \url{http://hera.inf-cv.uni-jena.de:6780}.

\section{Conclusions}
\label{sec:conclusions}

In this paper, we analyzed the influence of quantization in the bag-of-visual-words approach on the recognition performance of human
observers and compared it to the performance of an automatic visual recognition system. Throughout our analysis, we tried to establish a fair comparison
between human and machine performance as much as possible by providing each of them with the same local features. In particular, we inverted
quantized local features and presented them to observers in a human study, where the task was to perform scene recognition.

Our results showed that (i) humans perform significantly worse than machine learning approaches when being restricted to the visual 
information present in quantized local features rather than having access to the original input images, and 
(ii) that early stages of low level local feature extraction seem to be most crucial with respect to achieving human performance on original images. 
Finally, we demonstrated (iii) that large codebook sizes in the order of thousands of prototypes are essential not only for good machine learning performance, but more interestingly, also for human image understanding.

\bibliographystyle{IEEEtran}
\bibliography{tech-report}

\end{document}